\pdfoutput=1

\documentclass[11pt]{article}

\usepackage{acl}

\usepackage{times}
\usepackage{latexsym}
\usepackage{arydshln}
\usepackage[T1]{fontenc}

\usepackage[utf8]{inputenc}
\usepackage[breakable]{tcolorbox}
\usepackage{microtype}
\usepackage{amsfonts} 
\usepackage{pifont}
\usepackage{cancel}
\usepackage{soul}
\usepackage{xcolor}
\usepackage{algorithm}
\usepackage{algorithmic}
\usepackage{booktabs}
\usepackage{xspace}
\usepackage{mathrsfs}
\usepackage{multirow}
\usepackage{amsmath}
\usepackage{colortbl}
\usepackage{subfig}
\usepackage{graphicx}
\newcommand{\ie}{\textit{i.e.,}\xspace}

\definecolor{bleudefrance}{rgb}{0.19, 0.55, 0.91}
\definecolor{yes}{RGB}{239,211,69}
\definecolor{carminered}{rgb}{1.0, 0.0, 0.22}
\definecolor{crimsonglory}{rgb}{0.75, 0.0, 0.2}

\title{Flipping Knowledge Distillation: Leveraging Small Models' Expertise to Enhance LLMs in Text Matching}

\author{Mingzhe Li\textsuperscript{\rm 1}, Jing Xiang\textsuperscript{\rm 1}, Qishen Zhang\textsuperscript{\rm 1}, Kaiyang Wan\textsuperscript{\rm 2},
{Xiuying Chen\textsuperscript{\rm 2,}\thanks{\ \ \ Corresponding author.}}\ \\
$^1$ ByteDance, Beijing, China\\
$^2$ Mohamed bin Zayed University of Artificial Intelligence (MBZUAI)\\
\texttt{\{limingzhe.lmz, xiangjing, zhangqishen\}@bytedance.com},\\
\texttt{\{kaiyang.wan, xiuying.chen\}@mbzuai.ac.ae}\\
}

\begin{document}
\maketitle
\begin{abstract}
Knowledge distillation typically involves transferring knowledge from a Large Language Model (LLM) to a Smaller Language Model (SLM). 
However, in tasks such as text matching, fine-tuned smaller models often yield more effective domain-specific representations, as they focus on optimizing the similarity of input pairs. 
To leverage both the specialized strengths of small models and the rich semantic understanding of LLMs, we introduce a flipped knowledge distillation paradigm, where LLM learns from SLM.
Specifically, we address the architectural gap between decoder-only LLMs and smaller encoder-based models by reinterpreting LLMs in an encoder-decoder manner using LoRA. 
The encoder generates compressed representations, while the decoder maps them to the output space. 
During training, the encoder produces representations and their similarities, which are then aligned with the similarity scores produced by the teacher, using our proposed Margin-aware Contrastive Learning (MCL) approach. 
The MCL ensures accurate similarity for both positive and negative pairs, and adaptively handles the internal differences within positive and negative samples.
Our paradigm requires only a reasonably good-performing SLM, allowing the LLM to achieve improved performance. 
Experiments on financial and healthcare benchmarks, as well as real-world applications, confirm its effectiveness, and the model has been fully deployed in an online environment.
\end{abstract}

\section{Introduction}

Large Language Model (LLMs) have demonstrated remarkable capabilities in acquiring diverse knowledge, making them highly effective across a wide range of tasks \cite{zhao2023survey,xi2023rise,song2025injecting}. 
Consequently, Smaller Language Models (SLM) often learn from LLMs via knowledge distillation and imitation learning~\cite{gu2023knowledge,li2023distilling,gu2024minillm,xu2024survey}. 
However, despite their extensive knowledge, LLMs often underperform on domain-specific tasks compared to smaller models fine-tuned on specialized data \cite{ma2023large,chen2023robust,stewart2023large,hu2024bad}. 
For example, in text matching, SLMs are trained to make paired inputs more similar in representation~\cite{devlin2018bert}, unlike LLMs that directly predict match or non-match~\cite{touvron2023Llama}. 
This allows SLMs to better distinguish between synonyms and enhances their representation learning~\cite{hillebrand2023improving}, especially in specialized tasks, because they can learn in-domain terminology more effectively.

\begin{figure*}[t]
\centering
\includegraphics[scale=0.53]{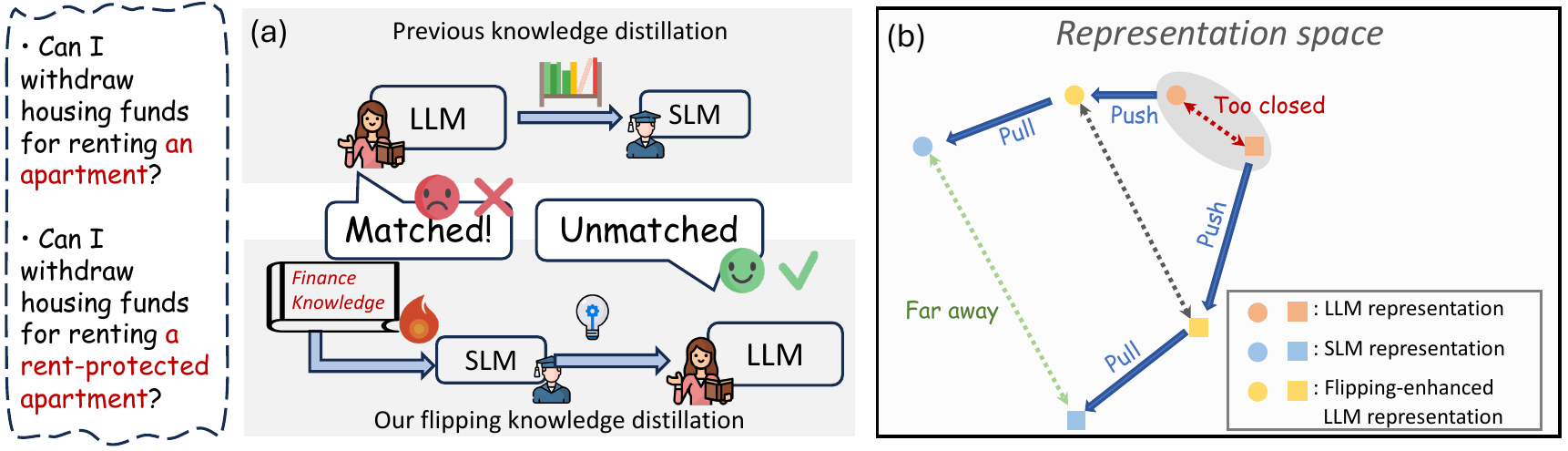}
\caption{
(a) Comparison of traditional knowledge distillation from LLM to SLM, and our flipping knowledge distillation of domain-specific expertise from SLM to LLM.
(b) In the representation space, when texts are highly similar but semantically unrelated, the LLM learns the correct semantic distance between them through distillation from the SLM, which provides more accurate knowledge.
}
\label{fig:intro}
\end{figure*}

Therefore, in this work, we propose a novel flipping knowledge distillation approach for LLMs to learn from the expertise of SLMs in representation learning for text matching, as illustrated in Figure~\ref{fig:intro}. 
Previous efforts to improve the performance of large models in specific domains predominantly focused on directly utilizing domain-specific data by supervised fine-tuning~\cite{ding2023parameter}, retrieval-augmented generation~\cite{zakka2024almanac}, or modifying model architectures to inject domain-specific knowledge~\cite{diao2023mixture,zhang2023plug}. 
In contrast, our method takes advantage of LLM's rich semantic understanding ability, while also enabling LLM to learn directly from the specialized knowledge of SLM.
This enables LLM to acquire not only simple matching labels, but also more nuanced representation information from SLM, thereby capturing more detailed knowledge.

Flipped knowledge learning presents two primary obstacles. 
First, state-of-the-art LLMs are predominantly decoder-only architectures, lacking dedicated modules for learning effective text representations. 
Second, effectively learning relationships and relative distances between and within positive and negative pairs from small model is difficult.
To address these challenges, we reinterpret the decoder-only LLM structure as an encoder-decoder mechanism by leveraging Low-Rank Adaptation (LoRA)~\cite{hulora}. 
In our approach, the encoder, represented by the compressed matrices in LoRA, learns compact input representations, while the decoder, defined by the expansion matrices, decodes and recombines these representations. 
This design allows the encoder to generate text representations and compute their similarities, which are then aligned with the similarity scores produced by the teacher model, providing guidance for capturing semantic similarity more effectively.
To further enhance learning from the teacher model, we propose a Margin-aware Contrastive Loss (MCL). 
Unlike conventional methods that only ensure positive samples have higher similarity scores than negative ones, MCL introduces two margin zones.
These zones encourage the LLM to learn greater differentiation not only between positive and negative samples but also within each category.
Additionally, to address potential inaccuracies in the teacher model, we introduce a dual threshold strategy to filter noises, ensuring more reliable learning.

For experiments, we evaluate our paradigm on various LLMs with different scale architectures, including Qwen-0.5b and GLM-10b, learning from three different SLMs across three benchmarks. 
The results demonstrate that our model not only outperforms the original LLMs but also surpasses LLMs trained using other strategies, such as supervised finetuning, parameter-efficient fine-tuning, and distillation from larger or smaller teacher models.
 
Our main contributions are as follows:
First, we introduce a paradigm that combines the strengths of LLMs and SLMs, incorporating domain-specific expertise from SLMs into LLMs, thus enhancing overall performance.
Second, we develop a dual margin learning function for knowledge distillation that effectively identifies and differentiates between positive and negative samples, improving distillation precision.
Finally, our approach yields a fine-tuned text matching model that outperforms existing methods, providing a robust solution for knowledge distillation.

\begin{figure*}[t]
\centering
\includegraphics[scale=1.14]{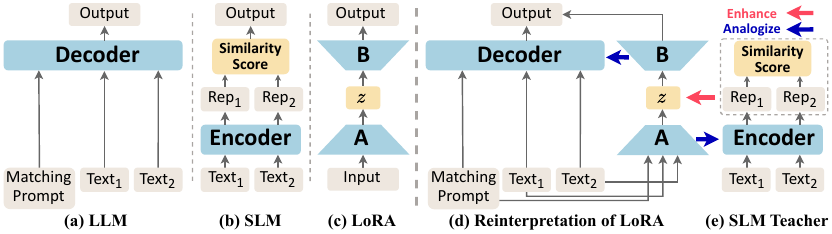}
\caption{
    Comparison of SLM, traditional LLM, and our reinterpreted LLM for text matching. 
    We reinterpret the decoder-only structure of the LLM into an encoder-decoder architecture to facilitate calculating the matching score between text representations instead of directly generating output.
}
\label{fig:rellm}
\end{figure*}

\section{Related Work}

\paragraph{Knowledge Distillation.}
Knowledge distillation is a task that involves transferring knowledge from an advanced model to a less performant model.
Distillation algorithms can achieve this in various ways. 
For example, a teacher model can generate data for the smaller model to learn from~\cite{chiang2023vicuna,xu2023wizardlm}.
One instance is Alpaca-7b~\cite{taori2023stanford}, a model fine-tuned from the Llama-7b model on 52,000 instruction-following demonstration.
Alternatively, the teacher model can produce features for the smaller model to learn~\cite{timiryasov2023baby}. 
For instance, \citet{liang2023less} propose task-aware layer-wise distillation to selectively capture relevant information from the teacher model using task-aware filters.
Other approaches include reinforcement learning \cite{chen2024improving} and ranking learning \cite{yuan2023rrhf}.
These works typically use a large teacher model (e.g., GPT-4, Llama-70b) and a small student model (e.g., GPT-2, Llama-7b).

\paragraph{Knowledge Injection.}
Before the era of LLMs, many works focused on manipulating the transformer structure to better inject knowledge.
For example, \citet{diao2023mixture} propose domain-specific adapters to inject domain-specific knowledge into the feedforward layer.  
\citet{zhang2023plug} introduce a knowledge plugin that injects knowledge  into frozen downstream models.
However, modifying the LLM architecture is challenging, and many researchers have instead turned to parameter-efficient fine-tuning (PEFT)~\cite{hulora,yang2023fingpt,chen2024flexible}, injecting knowledge via in-context learning \cite{zhou2024enhancing,shankar2024context}, or using retrieval-augmented generation \cite{xu2024unsupervised,wang2024m}. 
Since the knowledge that can be injected into the input is limited, we propose a new PEFT approach from a distillation perspective.

\paragraph{Text Matching.} 
Text matching is critically important in numerous applications.
For instance, in search engines, the semantic meaning of queries is matched with that of the returned websites \cite{li2024multi}.
In e-commerce, accurately matching queries to relevant products is essential for the success of these platforms \cite{fan2022modeling}. 
Another application involves rewriting user queries into pre-selected, high-performing queries to improve search outcomes \cite{chen2019rpm}. 
Most prior research has relied on smaller language models. 
For example, \citet{zou2022divide} proposed a divide-and-conquer approach for BERT by separating keywords from intents. 
\citet{chen2025unveiling} adopt BERT-based text matching for the chemical domain.
In this study, we investigate methods to evaluate and enhance the performance of LLMs in text matching tasks, aiming to take advantage of its internal ability.

\section{Method}

\subsection{Problem Formulation}
We begin by introducing the notations and key concepts. 
Formally, the text-matching task is to predict \( y_{i,j} \) for \( \{(x_i, x_j)\} \), where \( x_i \) and \( x_j \) represent the \(i\)-th and \(j\)-th input text pair, and \( y_{i,j} \) is the corresponding label, indicating whether the pair is a match.
Our objective is to enhance the performance of the student model \( \mathcal{S} \) in text matching by leveraging both the teacher model \( \mathcal{T} \) and the matching dataset \( D = \{(x_i, x_j, y_{i,j})\}_{i,j=1}^N \). 
To achieve this, we minimize a combined loss function, consisting of the supervised loss \( \mathcal{L}_{sup} \) on the training set \( D \),  a distillation loss \( \mathcal{L}_{dist} \) and a Margin-aware Contrastive Loss \( \mathcal{L}_{MCL} \), which capture the discrepancy between the predictions of \( \mathcal{T} \) and \( \mathcal{S} \).

In our setup, the student model is a LLM pre-trained on general domain and diverse datasets, while the teacher model is a smaller, specialized text-matching model.

\begin{figure*}[t]
\centering
\includegraphics[scale=0.43]{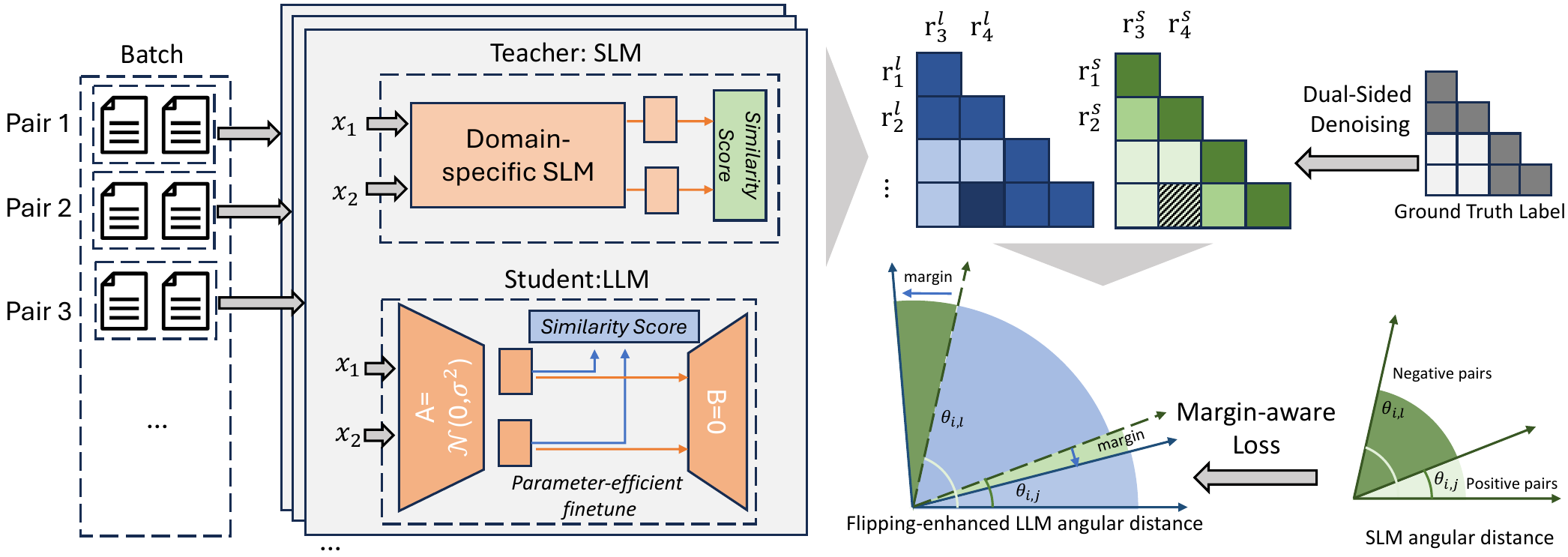}
\caption{
  Overview of our framework. The SLM and LLM output the representations for each text, forming a matrix. For distillation learning, we first use dual-sided denoising to filter out noisy outputs and design a margin-aware contrastive loss, which not only differentiates between positive and negative pairs but also regulates intra-group similarity within both positive and negative samples.
}
\label{fig:model}
\end{figure*}

\subsection{Reinterpretation of LLM}


Figure \ref{fig:rellm} illustrates various architectures for text-matching tasks and our proposed approach.

Figure \ref{fig:rellm}(a) shows that decoder-only LLMs process the task by taking a concatenated input of a matching prompt, $\text{Text}_1$, and  $\text{Text}_2$. 
However, this approach lacks explicit representation of input texts. 
In comparison, Figure \ref{fig:rellm}(b) demonstrates that encoder-only SLMs transform inputs into high-dimensional embeddings and are trained to compute their similarity to obtain a matching score. 
This design inherently captures explicit feature correlations between texts, enabling better modeling of pairwise relationships.

To bridge this architectural gap and enable LLMs to benefit from SLM-like approaches, we draw inspiration from LoRA and reinterpret the LLM as an encoder-decoder structure. 
In LoRA, a pre-trained weight matrix  $\mathbf{W}_0 \in \mathbb{R}^{d \times k}$  is updated using a low-rank decomposition  $\mathbf{W}_0 + \Delta \mathbf{W} = \mathbf{W}_0 + \mathbf{B} \mathbf{A}$, where  $\mathbf{B} \in \mathbb{R}^{d \times r}$  and  $\mathbf{A} \in \mathbb{R}^{r \times k}$, with  $r \ll \min(d, k)$.
During training, \( \mathbf{W}_0 \) is frozen and does not receive gradient updates, while \( \mathbf{A} \) and \( \mathbf{B} \) contain the trainable parameters.
Given \( \mathbf{h} = \mathbf{W}_0 \mathbf{x} \), the modified forward pass becomes:
\begin{align}
    \mathbf{h}' = \mathbf{W}_0 \mathbf{x} + \Delta \mathbf{W} \mathbf{x} = \mathbf{W}_0 \mathbf{x} + \mathbf{B} \mathbf{A} \mathbf{x}.
\end{align}
Here,  $A$  serves as an encoder-like transformation, mapping  \( \mathbf{x} \in \mathbb{R}^k\)  into a lower-dimensional representation  $\mathbf{z} \in \mathbb{R}^r$, while  $B$  acts as a decoder-like transformation, projecting  $\mathbf{z}$  back into output space as Figure \ref{fig:rellm}(c) shows.


Building on this insight, as shown in Figure~\ref{fig:rellm}(d) and \ref{fig:rellm}(e), we reinterpret LoRA as an encoder-decoder framework to integrate SLMs into LLMs effectively. 
Specifically, SLMs compute explicit similarity scores between  $\text{text}_1$  and  $\text{text}_2$, which serve as a teacher signal to enhance the relational reasoning capabilities of LLMs. 
By combining the strengths of both architectures, our proposed method establishes a robust synergy that improves performance on text-matching tasks.


\subsection{Knowledge Distillation Architecture}

Based on the above interpretation, we are able to set the LLM as the student model, learning from the SLM teacher model.
Concretely, assume the embeddings of the two input texts \( x_i \) and \( x_j \) are \( \mathbf{x}_i \in \mathbb{R}^{m \times k} \) and \( \mathbf{x}_j \in \mathbb{R}^{n \times k} \), where \( m \) and \( n \) represent the number of tokens. 
These embeddings are separately fed into the SLM, which could be a domain-specific language model like BERT.
By taking the average of the word-level vectors from the output, we obtain \( \mathbf{r}^s_i \) and \( \mathbf{r}^s_j \in \mathbb{R}^r \).

For the student LLM, the inputs are concatenated into a single input $\mathbf{x}^{l}$.
This concatenated input \( \mathbf{x}^{l} \) is then processed by the LoRA matrix \( \mathbf{A} \), which compresses the representation from the original $k$-dimensional space to a lower-dimensional subspace:
$\mathbf{r}^l =  \mathbf{x}^{l}\mathbf{A^T} \in \mathbb{R}^{(m+n) \times r}$, where the first  $m$  dimensions of  $\mathbf{r}^l$  are averaged to obtain  $\mathbf{r}^l_i$ , and the last  $n$  dimensions are averaged to obtain  $\mathbf{r}^l_j$.
Note that the compressed vector dimension in the LLM is set to match the original vector size \( r \) in the SLM for proper alignment.
Then $\mathbf{r}^l$ is further decoded by $\mathbf{r}^l\mathbf{B}^T$.
Finally, the LLM generates logits  $\mathbf{a} = [a_{\text{yes}}, a_{\text{no}}]$  for the prediction. 
These logits are passed through a softmax function to compute probabilities for each class as: $p_{yes} = \frac{e^{a_{yes}}}{\sum_{c{\prime} \in \{\text{yes}, \text{no}\}} e^{a_{c{\prime}}}}$.
The classification label is then optimized using the binary cross-entropy loss defined as: $\mathcal{L}_{sup} = - \left( y \log(p_{yes}) + (1 - y) \log(1 - p_{yes}) \right)$, where  $y$  is the ground truth label.

For distillation learning, an intuitive approach for the student model to learn from the teacher is to make the student's representation \( \mathbf{r}^l \) closer to the teacher's representation \( \mathbf{r}^s \). 
However, our goal is to learn the \textit{matching relationships} between documents, rather than replicating the original vector from the SLM. 
This motivates us to construct a pairwise matrix that stores these relationships and allows the model to learn the matching patterns, as shown in Figure~\ref{fig:model}.

\textbf{Threshold-aware Matching Matrix.}
The matching relationship matrix includes not only matched pairs but also unmatched inputs to enhance the robustness of learning. To handle negative examples, we combine the other inputs in the batch with the original input to form negative pairs. We calculate the cosine similarity for both the teacher (SLM) and student (LLM) models using the following equation:
\begin{align}
\alpha^m_{i,j} = \frac{\mathbf{r}^m_i \cdot \mathbf{r}^m_j}{\|\mathbf{r}^m_i\|_2 \|\mathbf{r}^m_j\|_2}, \quad m \in \{s, l\},
\end{align}
where \( \mathbf{r}^m_i \) and \( \mathbf{r}^m_j \) are vectors from the student model \( (l) \) or teacher model \( (s) \), and \( \alpha^m_{i,j} \) represents the cosine similarity between these two vectors.

To filter out less confident predictions from the teacher model, we define a threshold-based filtering function to remove unreliable negative samples from the teacher’s similarity matrix. Specifically, we use the cosine similarity \( \alpha^s_{i,j} \), compare it against a threshold \( \theta \), and also take into account the ground truth label \( y_{i,j} \), where \( y_{i,j} = 1 \) indicates a related pair, and \( y_{i,j} = 0 \) indicates an unrelated pair:
\begin{align}
    \small
\varphi^s_{i,j} = 
\begin{cases}
0, & \text{if } \left(\alpha^s_{i,j} < \theta \, \land \, y_{i,j} = 1\right) \\
   & \quad \text{or } \left(\alpha^s_{i,j} \geq 1-\theta \, \land \, y_{i,j} = 0\right), \\
1, & \text{otherwise}.
\end{cases}
\end{align}
The key idea here is that if a pair is labeled as related (\( y_{i,j} = 1 \)) but has a cosine similarity \( \alpha^m_{i,j} \) below the threshold \( \theta \), or if a pair is labeled as unrelated (\( y_{i,j} = 0 \)) but has a similarity above \( 1-\theta \), we consider these pairs to be noisy and filter them out. This ensures that only consistent and confident pairs contribute to the distillation loss.

Finally, we define the distillation loss with threshold filtering as below:
\begin{align}
    \mathcal{L}_{\text{dist}} = \frac{1}{n} \textstyle \sum_{i,j} \varphi^s_{i,j} \left( \alpha^s_{i,j} - \alpha^l_{i,j} \right)^2.
\end{align}


\subsection{Margin-aware Contrastive Learning}
The matrix loss described earlier captures the pairwise similarity between the teacher and student models but does not \textit{explicitly enforce contrast between related and unrelated pairs}.
Moreover, it does not account for \textit{differences within the positive and negative pairs}. 
To address this, we propose a margin-aware contrastive loss, which introduces a margin to regulate intra-group similarity within both positive and negative samples.

We begin by transforming the cosine similarity of the student model into angular distance using the arccosine function. 
This transformation enhances the model’s discriminative power by amplifying subtle differences in similarity, making it easier to distinguish between similar and dissimilar samples~\cite{zhang2022contrastive}:
\begin{align}
\theta^s_{i,j} = \arccos(\alpha^s_{i,j}),
\theta^l_{i,j} = \arccos(\alpha^l_{i,j}).
\end{align}
We then introduce a contrastive loss $\mathcal{L}_{\text{MCL}}$ between the teacher and student models:
\[
\resizebox{0.5\textwidth}{!}{$\mathcal{L}_{\text{MCL}} = -\log \frac{\varphi^s_{i,j}e^{\cos(\theta^l_{i,j} + m_c\theta^s_{i,j})}}{\varphi^s_{i,j}e^{\cos(\theta^l_{i,j} + m_c\theta^s_{i,j})} + \sum_{i,j'} \varphi^s_{i,j'}e^{\cos(\theta^l_{i,j'}- m_c\theta^s_{i,j})}}$}
\]
where \( m_c \) is a scalar controlling the margin influence from the teacher model. 
 $i, j$  are paired samples, while  $i, j'$ are unpaired samples.

Overall, contrastive learning helps the model distinguish between positive and negative pairs. 
Additionally, the contrastive degree is influenced by the teacher model's angular distance, \( \theta^s_{i,j} \).
For positive pairs, a margin is added to the angular distance, encouraging the student model to produce smaller angular distances for pairs with higher teacher similarity scores.
Conversely, pairs with lower teacher-assigned similarity scores are assigned larger margins, leading to relatively lower (but still positive) similarity scores.
For negative pairs, the margin is subtracted, increasing the angular distance and reducing the similarity scores.
In both cases, the margin space allows the LLM to effectively learn relationships between positive and negative pairs, as well as the relative distances within pairs, from the small model's output representations.

Together, the LLM is trained using the $\mathcal{L}_{\text{dist}}$ and $\mathcal{L}_{\text{MCL}}$ losses from the teacher model while simultaneously learning from the dataset through the supervised loss $\mathcal{L}_{\text{sup}}$.

\section{Experiment}
\label{sec:experiment}

\subsection{Dataset}
To comprehensively evaluate the model, we select two domain-specific text matching datasets and one real-world online dataset from ByteDance.

C-MTEB \cite{xiao2023c} is a benchmark for Chinese text embeddings, covering 6 tasks and 35 datasets. 
We use the \textbf{ATEC} dataset within C-MTEB, which is designed to determine whether two sentences in the financial domain are related. 
\textbf{NFCorpus} \cite{boteva2016} is an English medical information retrieval dataset.
The task is to determine the relevance between a title and a document. 
The third dataset is our in-house dataset from \textbf{ByteDance}, which covers multiple product lines, such as payment services, subscription services, insurance, and personal loan services, sourced from real online customer service interactions. 
The task is Frequently Asked Question (FAQ) retrieval, which determines whether a user’s query is relevant to another well-written and frequently asked question in the pre-prepared dataset.
The matching labels are manually annotated. 
The training dataset contains 113,316 examples, with 10,000 examples in both the development and test sets. 
More details can be found in Appendix \ref{appendix:dataset}.

\subsection{Comparison Methods}

For teacher models, we select the following \textit{SLMs} tailored for specific domains: GTE~\cite{li2023towards}, an embedding model trained with multi-stage contrastive learning for multi-domain; FinBERT~\cite{yang2020finbert}, a BERT model fine-tuned specifically for the finance domain; and MedBERT~\cite{9980157}, fine-tuned for the medical domain.
For the student LLM backbone, we use Qwen-0.5b~\cite{bai2023qwen} and GLM-10b~\cite{glm2024chatglm}.

In terms of distillation methods, we include \textit{LLMs that learn from larger LLMs}, including Upaya~\cite{jindal2024upaya} and PMC-Llama~\cite{wu2024pmc}.
Upaya is distilled from Llama-70b to Llama-7b for the finance domain, while PMC-Llama is distilled from ChatGPT for the medical domain.
We also include \textit{LLMs learned from SLMs} with baseline-version flipped distillation, where the reinterpretation of the LLM is omitted, and the LLM learns solely from the similarity scores generated by the SLM using a carefully designed loss function.
This comparison includes ArcCSE~\cite{zhang2022contrastive}, which learns sentence representations in angular space, and KDMCSE~\cite{nguyen2024kdmcse}, which uses adaptive angular margin contrastive learning to enhance discriminative representations while capturing negative semantics.


\begin{table*}[t]
    \centering
    \small
    \resizebox{\textwidth}{!}{%
    \begin{tabular}{@{}l|ccc|ccc|ccc@{}}
      \toprule
      & \multicolumn{3}{c|}{ATEC} &  \multicolumn{3}{c}{Nfcorpus} &  \multicolumn{3}{c}{Bytedance} \\
      Model & Acc & F1 & AUC & Acc & F1 & AUC & Acc & F1 & AUC \\
    \midrule
      \multicolumn{9}{@{}l}{\textit{SLM:}} \\
      GTE & 0.8411 & 0.8003 & 0.8348 & 0.9402 & 0.8790 & 0.9277 & 0.7794 & 0.7005 & 0.9010\\
      FinBERT & 0.8453 & 0.8025 & 0.8369 & 0.9371 & 0.8714 & 0.9232 & 0.7782 & 0.7017 & 0.9021\\
      MedBert & 0.8388 & 0.7979 & 0.8306 & 0.9375 & 0.8847 & 0.9324 & 0.7740 & 0.7004 & 0.8971\\
      \midrule
       \multicolumn{9}{@{}l}{\textit{LLM:}} \\
      Qwen-0.5b & 0.8623 & 0.8133 & 0.8567 & 0.9593 & 0.8947 & 0.9481 & 0.7968 & 0.7206 & 0.9140\\
      Qwen-0.5b-LoRA  & 0.8617 & 0.8151 & 0.8560 & 0.9588 & 0.8963 & 0.9440 & 0.7912 & 0.7144 & 0.9175 \\
      Llama-7b & 0.8863 & 0.8307 & 0.8818 & 0.9679 & 0.9066 & 0.9591 & 0.8093 & 0.7297 & 0.9123\\
      GLM-10b & 0.8869 & 0.8351 & 0.8826 & 0.9774 & 0.9107 & 0.9673 & 0.8111 & 0.7308 & 0.9315 \\
      GLM-10b-LoRA  & 0.8890 & 0.8322 & 0.8799 & 0.9717 & 0.9068 & 0.9604 & 0.8117 & 0.7271 & 0.9289 \\
            \midrule
        \multicolumn{9}{@{}l}{\textit{\textit{LLM w/ classic distillation:}}} \\
      Upaya (Llama-7b)/PMC (Llama-13b)
      & 0.8934 & 0.8378 & 0.8895 & 0.9843 & 0.9227 & 0.9756 & - & - & - \\
      \midrule
       \multicolumn{9}{@{}l}{\textit{\textit{LLM w/ baseline-version flip distillation:}}} \\
        Qwen-0.5b (ArcCSE) & 0.8689 & 0.8162 & 0.8631 & 0.9621 & 0.8949 & 0.9504 & 0.8008 & 0.7235 & 0.9147 \\
      Qwen-0.5b (KDMCSE) & 0.8727 & 0.8199 & 0.8668 & 0.9638 & 0.9025 & 0.9582  & 0.8017 & 0.7254 & 0.9160 \\
        \midrule
          \multicolumn{9}{@{}l}{\textit{\textit{LLM w/ our flip distillation:}}}\\
      Qwen-0.5b-flip (FinBERT/MedBERT) & \textbf{0.8784} & \textbf{0.8247} & 0.8705 & \textbf{0.9707} & \textbf{0.9132} & \textbf{0.9710}& - & - & - \\
      Qwen-0.5b-flip (GTE) & 0.8746 & 0.8203 & \textbf{0.8719} & 0.9703 & 0.9092 & 0.9638 & \textbf{0.8076} & \textbf{0.7307} & \textbf{0.9251} \\
      GLM-10b-flip (FinBERT/MedBERT) & 0.8937 & \textbf{0.8449} & \textbf{0.8914} & \textbf{0.9866} & \textbf{0.9249} & \textbf{0.9774} & - & - & - \\
      GLM-10b-flip (GTE) & \textbf{0.8963} & 0.8422 & 0.8869 & 0.9791 & 0.9204 & 0.9759 & \textbf{0.8248} & \textbf{0.7415} & \textbf{0.9406} \\
         \midrule
        {\textit{Ablation Study of our Qwen-0.5b-flip (GTE):}} \\
      Qwen-0.5b-flip (GTE) w/o $\mathcal{L}_{\text{MCL}}$ & 0.8631 & 0.8154 & 0.8593 & 0.9668 & 0.9025 & 0.9637 & 0.7982 & 0.7226 & 0.9169 \\
       Qwen-0.5b-flip  (GTE) w/o $\mathcal{L}_{\text{dist}}$ & 0.8654 & 0.8147 & 0.8593 & 0.9656 & 0.9022 & 0.9670 & 0.7991 & 0.7249 & 0.9180 \\
       Qwen-0.5b-flip (GTE) w/o noise filtering & 0.8662 & 0.8177 & 0.8659 & 0.9683 & 0.9065 & 0.9601 & 0.8014 & 0.7252 & 0.9203 \\
      \bottomrule
    \end{tabular}}
    \caption{Offline performance on three datasets. 
    Rows like "Upaya (Llama-7b)/PMC (Llama-13b)" show Upaya’s performance on ATEC and PMC’s performance on NfCorpus.
    Numbers in \textbf{bold} mean that the improvement to the backbone model is statistically significant (a two-tailed paired t-test with p-value \textless 0.01).}
    \label{tab:exp}
  \end{table*}

\subsection{Implementation Details}
We implemented our experiments in PyTorch on NVIDIA H20 GPUs. For our model and all baselines, we followed the same settings. Qwen-0.5b models were trained on one GPU, and GLM-10b models on four GPUs. The truncation lengths for the ATEC, NFCorpus, and ByteDance datasets are 256, 2048, and 128, respectively. The same prompt was used for text matching across models. During LLM training, the LLM’s parameters were fixed, and LoRA was integrated into the attention mechanism of each layer, with its rank matching the teacher model’s hidden dimension to capture teacher representations. The hyperparameter $m_c$ was set to 0.06 (Section~\ref{sec:mcl}), and threshold $\theta$ to 0.5 for balanced binary classification. Loss functions $\mathcal{L}_{dist}$ and $\mathcal{L}_{MCL}$ were applied to the LoRA of the last layer, while $\mathcal{L}_{sup}$ was applied to the final output logit. The final loss was calculated by summing all $L_{*}$, weighted by their respective scales. We selected the 5 best checkpoints based on validation set performance and reported averaged test set results.
Our method significantly reduces computational requirements, training only 22\% of parameters for Qwen-0.5b and 7.54\% for GLM-10b.
More details can be found in the Appendix~\ref{appendix:implementation}.

\subsection{Evaluation Metrics}
For offline evaluation, we use Accuracy for correct predictions, AUC for class imbalance, and F1 for balancing precision and recall.
For online evaluation, we assess the performance of FAQ tasks using localization accuracy, which measures the alignment between user queries and returned search results.
To ensure practical applicability, our matching module is fully integrated with upstream and downstream components, allowing us to evaluate the system’s end-to-end performance.
Users provide feedback on whether the final search results align with their intended queries.
We compare the end-to-end user satisfaction of our matching module against the baseline model, Baichuan-7b.

\subsection{Offline Performance}

Table~\ref{tab:exp} presents the performance of diverse baselines and our model under various settings.

Firstly, \textit{SLMs and LLMs pre-trained in the same domain exhibit better performance}. 
For instance, FinBERT, fine-tuned in the financial domain, achieves better performance than the other two SLMs on ATEC, with an F1 score of 0.8025 compared to GTE’s 0.8003 and MedBERT’s 0.7979. 
Similarly, MedBERT outperforms GTE and FinBERT on NfCorpus. These results emphasize the importance of domain-specific pretraining in achieving better task-specific performance.
Secondly, \textit{distillation from both larger LLMs and smaller SLMs significantly boosts performance}. 
For example, Upaya (Llama-7b), distilled from a larger LLM, demonstrates a substantial improvement over Llama-7b on ATEC, with the AUC rising from 0.8818 to 0.8895. 
These findings validate the effectiveness of leveraging both larger and smaller models during distillation to transfer domain-specific expertise.

Finally, \textit{our proposed flip distillation strategy performs best across datasets, architectures, and scales. }
For instance, FLM-10b-flip (MedBERT) with only 10 billion parameters achieves an F1 score of 0.9249 on Nfcorpus, surpassing PMC-Llama (Llama-13b), which achieves an F1 score of 0.9227 on NfCorpus, despite PMC-Llama being significantly larger. 
Furthermore, this improvement is consistent across models of varying scales, architectures, and domains. 
It is worth noting that even when the teacher model underperforms the student model, our method effectively facilitates knowledge transfer to enhance the student’s capabilities.
As shown in Table~\ref{tab:exp}, using the GTE model as the teacher for the Qwen-0.5b student model, our framework improved the student to achieve 0.8076 in accuracy.
These results highlight the generalizability and robustness of our flip distillation. 

\subsection{Online Experiments}

\begin{table}[t]
\small
\centering
\resizebox{\columnwidth}{!}{%
    \begin{tabular}{@{}l|cccc@{}}
      \toprule
      & \multicolumn{4}{c}{Scene} \\
      Method & Insurance & Loan & Payment & Subscription \\
      \midrule
      Qwen-0.5b-flip & \textbf{+10.64\%} & \textbf{+10.08\%} & \textbf{+1.05\%} & \textbf{+5.30\%} \\
      \bottomrule
    \end{tabular}%
}
\caption{The performance improvements on ByteDance product with online A/B test.}
\label{tab:online_exp}
\end{table}

Besides the offline experiments, we also conducted online A/B tests by deploying the flip distillation method in the FAQ task of the ByteDance search system over seven days. 
The control group utilized the matching strategy deployed in the current online system, specifically the Baichuan-7b model, which serves as our baseline as it is the model currently in use.
The average A/B test results across different scenarios are presented in Table~\ref{tab:online_exp}, with all results reported as relative improvements. 
Our proposed method achieves significant performance gains across all scenarios, including insurance, loan, payment, and subscription, highlighting the effectiveness of flip distillation in improving search results and user satisfaction within ByteDance’s ecosystem.

\section{Analysis and Discussion}

\subsection{Ablation Study}

We conduct an ablation study to evaluate the impact of different components in our proposed model, as shown in Table~\ref{tab:exp}. 
Removing $\mathcal{L}_{MCL}$ resulted in a noticeable drop in performance across all metrics. 
On the ByteDance dataset, the F1 score decreased from 0.7307 to 0.7226, and the AUC dropped from 0.9251 to 0.9169, demonstrating the importance of the margin knowledge from teacher model.
When $\mathcal{L}_{dist}$ was removed, the performance also declined. 
For instance, the F1 score on ByteDance decreased from 0.7307 to 0.7249, and the AUC fell from 0.9251 to 0.9180.
Eliminating the noise filtering, which filters the noisy learning cases, resulted in the smallest performance drop compared to the other components. 
Overall, the results demonstrate the necessity of each component in achieving optimal performance, with $\mathcal{L}_{MCL}$ and $\mathcal{L}_{dist}$ playing crucial roles in maintaining the robustness and adaptability of the model across diverse datasets.

\subsection{Impact of Margin in $\mathcal{L}_{\text{MCL}}$}
\label{sec:mcl}

\begin{figure}[t]
\centering
\includegraphics[scale=0.23]{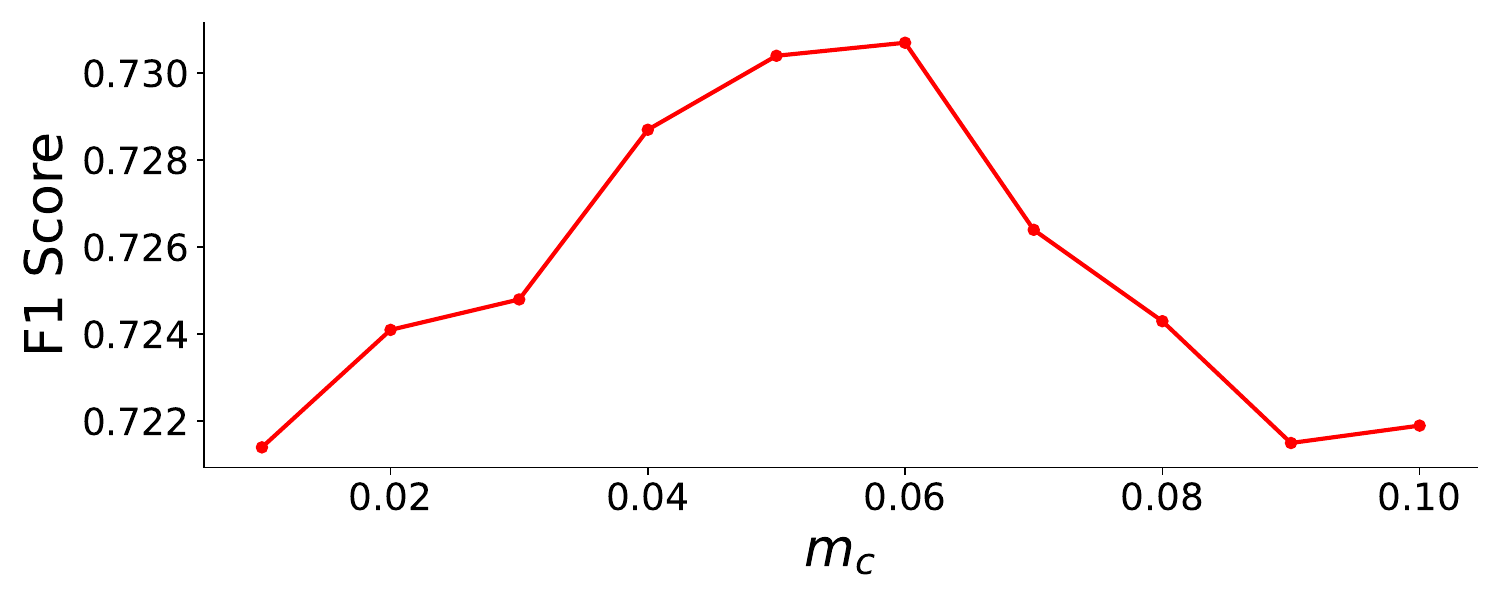}
\vspace{-2mm}
\caption{Impact of changing angular margin $m_c$ in $\mathcal{L}_{\text{MCL}}$. We take the F1 results of Qwen-0.5-flip (GTE) on Bytedance as example.}
\label{fig:margin}
\end{figure}

Figure~\ref{fig:margin} illustrates the effect of varying the margin  $m_c$ in the MCL loss function. 
It is evident that the angular margin significantly influences the model’s performance. 
The F1 score steadily increases as  $m_c$  grows from 0.0 to approximately 0.06, peaking at 0.7307. 
As  $m_c$  continues to increase beyond 0.06, the F1 score starts to decline, dropping to 0.7219 when  $m_c$  reaches 0.1.
The results align with our hypothesis that an overly small angular margin fails to fully utilize the margin-aware adaptability of MCL, resulting in suboptimal similarity alignment.
Conversely, an excessively large margin distorts the representation space, leading to degraded performance. 

\begin{figure}[t]
\centering
\includegraphics[scale=0.29]{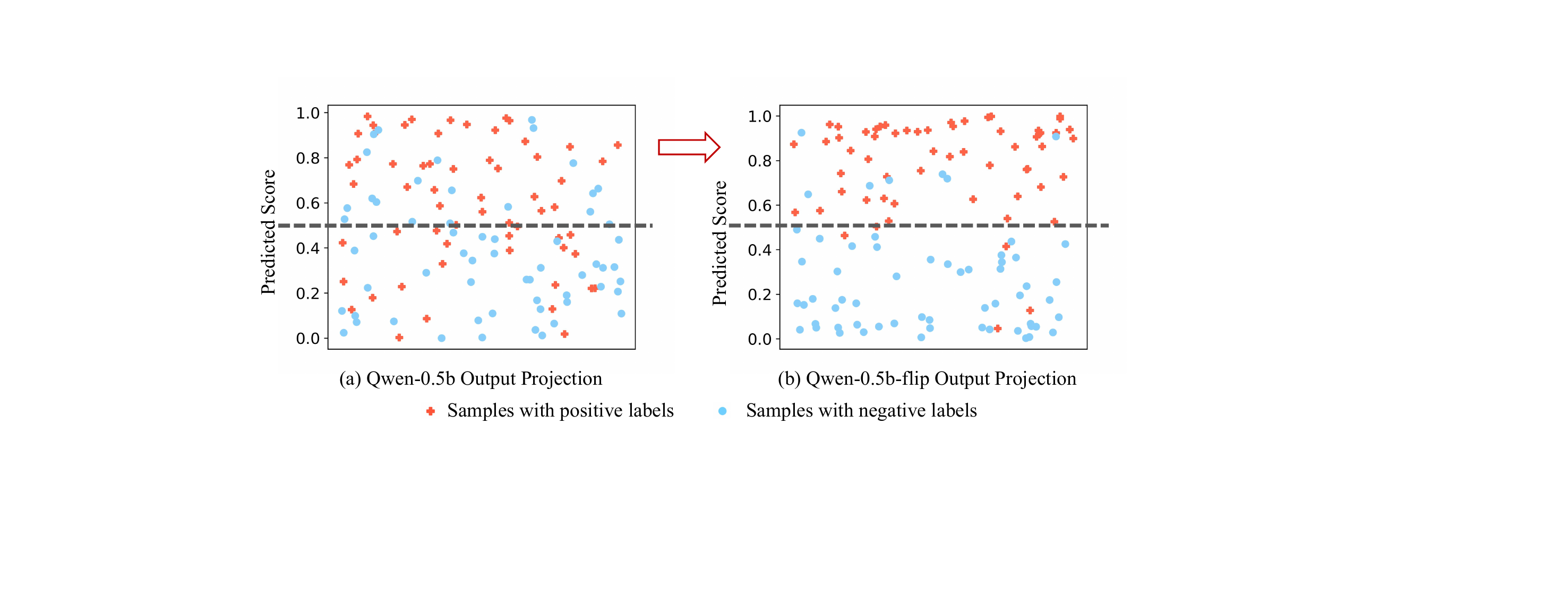}
\caption{Distribution of positive and negative samples from Qwen-0.5b and Qweb-0.5b-flip respectively, where our MCL loss brings a clear separation. 
}
\label{fig:point}
\end{figure}

\subsection{Visualization and Case Study}
Figure~\ref{fig:point} compares the distribution of samples predicted by Qwen-0.5b and Qwen-0.5b-flip. 
The left part shows overlap between positive and negative samples near the decision boundary, indicating poor separability.
In contrast, the right part shows clearer separation, with positive samples near 1.0 and negative samples near 0.0.
This flip distillation improves decision boundary clarity and class separability, aligning with higher F1 scores and better model performance.

\begin{figure}[htb]
\centering
\includegraphics[scale=0.4]{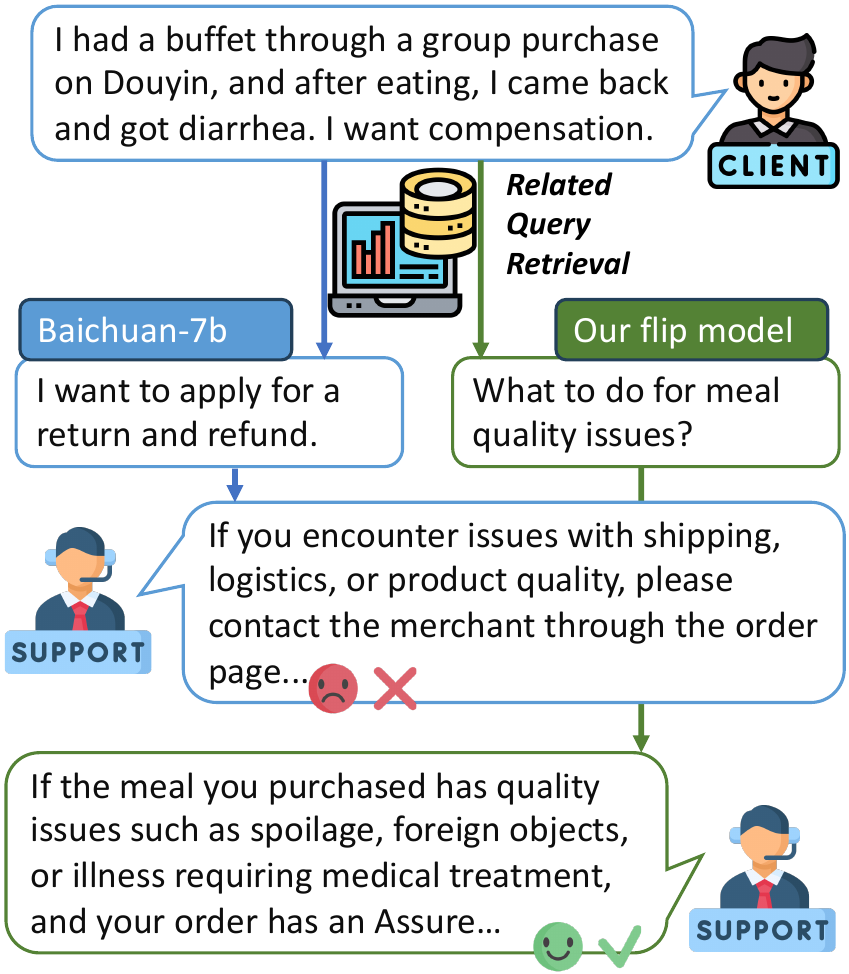}
\caption{Case study: Baichuan-7b retrieves irrelevant shipping info, while our method retrieves the correct question and gives an accurate response.
}
\label{fig:case}
\end{figure}

We also present a case study from the ByteDance dataset to demonstrate our method’s effectiveness in Figure~\ref{fig:case}. 
A user reports food poisoning after a buffet purchase on Douyin and requests compensation. 
The Baichuan-7b model retrieves an irrelevant response about shipping and logistics, failing to address the user’s issue. 
In contrast, our method retrieves the correct question, “What to do for meal quality issues?” and provides detailed instructions on filing a compensation claim through the “Assure Eat” program or contacting the lifestyle services hotline. 
This example highlights our method’s ability to align user intent with relevant responses, improving accuracy and user satisfaction.

\section{Conclusion}
In this paper, we propose a flipped distillation technique, where LLMs learn from smaller language models due to their superior representation learning ability in text matching tasks, reversing the traditional distillation direction.
Additionally, we introduce an adapted margin-aware distillation loss to enhance the learning process for LLMs. 
Experimental results across three domain-specific datasets and various model architectures consistently demonstrate the effectiveness of our approach.
In the future, we aim to further explore the potential of SLMs in a wider range of tasks to improve efficiency.

\section*{Limitation}
In this paper, we propose a flipped knowledge distillation paradigm where an LLM learns from an SLM to improve its performance on tasks such as text matching. 
However, one limitation is that our paradigm relies heavily on the high performance of the SLM. 
If an SLM does not perform well or fails to produce effective domain-specific representations, the knowledge transferred to the LLM may be suboptimal, leading to limited performance gains.
Another limitation is that our work focuses exclusively on the text matching task and does not explore its applicability to text generation tasks, which are widely relevant in many NLP applications. The current design and evaluation are tailored to tasks involving representation alignment and similarity computation.
Extending this paradigm to generation tasks, such as summarization or dialogue systems, would require additional considerations, including how knowledge alignment can influence text fluency and creativity. 
Future work could investigate the broader applicability of our approach across diverse NLP tasks to better understand its generalization capabilities.

\section*{Ethical Consideration}
However, several ethical considerations must be addressed.
First, bias in the smaller models could be transferred or amplified during distillation, requiring careful evaluation and mitigation. 
Second, the use of domain-specific data raises privacy concerns, necessitating strict data handling and anonymization practices. Third, while more efficient than traditional fine-tuning, LLM deployment still incurs notable environmental impact due to high computational costs.
Finally, for real-world applications in sensitive domains like healthcare and finance, rigorous testing is essential to ensure reliability and prevent harmful outcomes. Addressing these issues will promote responsible and fair deployment of AI systems.

\bibliography{custom}

\appendix
\section{Dataset}
\label{appendix:dataset}
C-MTEB \cite{xiao2023c} is a comprehensive benchmark for Chinese text embeddings, covering 6 tasks and 35 datasets. 
We use the \textbf{ATEC} dataset within C-MTEB, which is designed to determine whether two sentences in the financial domain are related. 
The training dataset has 100,000 examples, and both the development and test datasets have 2,477 examples each. The average sentence length is 26.79 words, with a maximum of 166 words and a minimum of 10 words.

\textbf{NFCorpus} \cite{boteva2016} is an English full-text retrieval dataset for Medical Information Retrieval. 
It contains 3,244 natural language queries, written in non-technical English, which were harvested from the NutritionFacts.org website. 
These queries are paired with 169,756 automatically extracted relevance judgments for 9,964 medical documents, mostly from PubMed. 
The task is to determine the relevance between a title and a document. 
The training dataset has 110,575 examples, with 11,385 in the development set and 12,334 in the test set. Titles average 17.61 words, and documents average 1,199.69 words.

The third dataset is our in-house dataset from \textbf{ByteDance}, which covers multiple product lines, such as payment services, subscription services, insurance, and personal loan services, sourced from real online customer service interactions. 
The task is Frequently Asked Question (FAQ) retrieval, which determines whether a user’s query is relevant to another well-written and frequently asked question in the pre-prepared dataset.
The matching labels are manually annotated. 
The training dataset contains 113,316 examples, with 10,000 examples in both the development and test sets. 
The average query length is 14.36 words, and the average length of FAQ questions is 9.73 words.

\section{Implementation Details}
\label{appendix:implementation}
We implemented our experiments in Pytorch on NVIDIA H20 GPU.
For our model and all baselines, we followed the same setting as described below. 
We trained the models based on Qwen-0.5b using one GPU at a time, while for models based on GLM-10b, we conducted experiments using 4 GPUs.
The truncation length for the ATEC, NFCorpus, and ByteDance datasets are 256, 2048, and 128, respectively.
The prompt for text matching in the LLM is set the same for both our model and the baselines as below:

ATEC prompt:
\begin{tcolorbox}[colback=gray!10, left=1mm, right=1mm, top=1mm, bottom=1mm] 
You are an expert in the financial domain. Based on your knowledge, determine if the semantic meaning of the following two sentences is the same. Sentence 1: “{sentence1}” and Sentence 2: “{sentence2}.” Please respond directly with “Yes” or “No.”
\end{tcolorbox}
NFC prompt:
\begin{tcolorbox}[colback=gray!10, left=1mm, right=1mm, top=1mm, bottom=1mm] 
You are an expert skilled in understanding medical content. Based on your professional knowledge, please determine if the following two items, titled "{sentence1}" and the article "{sentence2}", can be considered to describe the same content. Please answer directly with "Yes" or "No".
\end{tcolorbox}
Bytedance prompt:
\begin{tcolorbox}[colback=gray!10, left=1mm, right=1mm, top=1mm, bottom=1mm] 
You are an intelligent customer service assistant for financial products. Determine if the following two questions, Question 1: “{sentence1}” and Question 2: “{sentence2},” correspond to the same answer. Please respond directly with “Yes” or “No.”
\end{tcolorbox}
For LLM training, we kept the parameters of the LLM unchanged, and integrated LoRA into the attention mechanism of each layer. 
The parameters of LoRA were all initialized randomly using a zero-mean Gaussian distribution with a std of 0.01.
The rank and alpha of LoRA is determined based on the hidden dimension of the teacher model, in order to enable the model to learn the representations of the teacher model, \ie the rank of LoRA is equal to the dimension of the teacher model.
To enhance the robustness and generalization of the LoRA model, we set the dropout ratio as 0.05.
The hyperparameter $m_c$ is set to 0.06 according to our experiment in Section~\ref{sec:mcl} and threshold $\theta$ is 0.5, as it represents the balanced midpoint in binary classification tasks.
The $\mathcal{L}_{dist}$ and $\mathcal{L}_{MCL}$ are applied on the last layer of LLM.
\textcolor{black}{We calculate the final loss function as $Loss = L_{sup} + 0.1*L_{dist} + 0.1*L_{MCL}$. We choose these weights because we want to scale the components to the same magnitude.}
We set learning rate as 3e-5, and the warmup ratio as 0.05, which indicates the proportion of total training steps occupied by warmup.
Gradients were clipped to the range [-1, 1] during training to prevent exploding gradients.
We also added a weight decay of 0.1 to prevent overfitting, which is a regularization technique used during the training process.
\textcolor{black}{Our proposed LoRA-based framework significantly reduces the number of trainable parameters and computational requirements. For instance, with Qwen-0.5b as the student model, only 22\% of parameters are trained, while for GLM-10b, this reduces to 7.54\%.
This leads to notable time savings: in the Qwen-0.5b ByteDance experiment (batch size = 32), full-parameter SFT takes 5–6 hours, whereas LoRA completes training in just 2–3 hours. Regarding gradient calculation for our loss functions $L_{\text{dist}}$ and $L_{\text{MCL}}$, additional experiments showed that removing these loss terms did not significantly affect the overall training time.}
We selected the 5 best checkpoints based on performance on the validation set and reported averaged results on the test set.

\end{document}